# Towards the Design of Aerostat Wind Turbine Arrays through AI

Larry Bull & Neil Phillips

Department of Computer Science & Creative Technologies
University of the West of England, Bristol, UK

**Extended Abstract**

**Abstract** A new form of aerostat wind generation system which contains an array of interacting turbines is proposed. The design of the balloon turbine components is undertaken through the combination of artificial intelligence and rapid prototyping techniques such that the need for highly accurate models/simulations of the lift and wake dynamics is removed/reduced. Initial small-scale wind tunnel testing to determine design and algorithmic fundamentals will be presented.

As wind power makes an increasing contribution to the world's energy supply mix, there is growing interest in emerging technologies with potential niche advantages over the dominant tower-based horizontal axis turbines. Airborne wind generation systems represent a relatively unexplored source of renewable energy with a number of such advantages, including the exploitation of high-altitude winds onshore and offshore, temporary/mobile (significant) energy generation capabilities in developing countries or during disaster relief, a relatively low installation footprint and cost offshore, etc. These systems can be divided into two classes: kite systems, where the movement of a wing(s) harnesses the wind's energy; and aerostat systems, where a balloon is used to provide buoyancy to maintain the altitude of a turbine. Of the two approaches, the former has perhaps attracted the most attention thus far, as demonstrated by Google's acquisition of Makani Power in 2013 for example. Aerostats can be further sub-divided into two types: traditional wind turbine carrying systems and those where the balloon forms a part of the turbine. For example, Altaeros Energies' Buoyant Airborne Turbine consists of a central horizontal axis wind turbine surrounded by a balloon. Whereas Magenn Power's Air Rotor System is blimp-like but orientated sideways into the air flow, has turbine blades running along its length at evenly spaced intervals around its circumference, and is tethered at each end to a generator such that energy is produced as it rotates about its axis. The latter may be seen as a variant of a traditional vertical axis wind turbine laid horizontally. Horizontal axis wind turbines must be spaced apart to avoid wake interference whereas it has long been known that positive interactions can be generated between closely spaced vertical axis wind turbines (eg, [Charwat, 1978]). That is, an array of appropriately spaced vertical axis turbines can produce more energy than the sum of its individual turbines used in isolation. It is here proposed that similar levels of power to a single, large aerostat wind energy system will be possible with an array of smaller, interacting balloons that are both easier to operate and manufacture, with greater potential for scale-up. We are currently exploring examples of such aerostat systems, exploiting artificial intelligence (AI) and rapid prototyping techniques in the design process.

We are developing an approach to harness the creativity of artificial intelligence directly in physical space through the use of modern rapid prototyping techniques – what we've termed design mining [Preen & Bull, 2014]. In particular, it is intended for engineering design when simulations/models of sufficient accuracy do not exist, also enabling the exploitation of novel materials and manufacturing processes. The approach will also be of use for complex/novel design in general, eg, within the circular economy. The advent of additive-layer manufacturing (3D printing) has greatly enhanced the ability for designs to be rapidly created in a physical form from a wide variety of materials. As we have recently reviewed elsewhere [Preen & Bull, 2017], techniques such as evolutionary algorithms have been used in conjunction with rapid prototyping techniques but this has typically involved the physical realization of a design at the end of the optimisation process. In our approach the production of physical designs is inherent to the optimisation process. In the *simplest* case, design mining assumes no prior knowledge about the given domain and builds an initial model of the problem space through the testing of fabricated designs, whether specified by a human and/or machine. Optimisation techniques are then used to find the optima within the model of the collected data; the model which maps design specifications to performance is inverted and suggested good solutions identified. These are then fabricated and tested. The resulting data is added to the existing data and the process repeated. Over time the

model – built solely from physical prototypes tested appropriately for the task requirements – begins to capture the salient features of the problem, thereby enabling the discovery of high-quality (novel) solutions. Such so-called surrogate models have also long been used in optimisation for cases when multiple computational simulations are not viable due to their expense. We use them due to unknown or unquantifiable conditions (eg, [Bull, 1997]). Moreover, the overall task can be decomposed and each sub-task tackled concurrently for efficiency/scalability: the optimisation and modelling processes can be applied to individual elements of the overall task in parallel. The use of design mining will typically include consideration of the trade-off between model/simulation inaccuracies and their computational overheads with the cost of iterative physical solution creation and testing. Although, these need not be mutually exclusive since the data driven models created during the latter can be used to increase the accuracy of the former. That is, conducting optimisation directly through 3D printing-enabled experimentation and data mining opens new routes to the design of physical objects for complex problems. We have demonstrated that this can be more efficient than running the given optimisation process without the surrogate model(s) such that every proposed solution is simply realized physically and tested (eg, [Preen & Bull, 2016]).

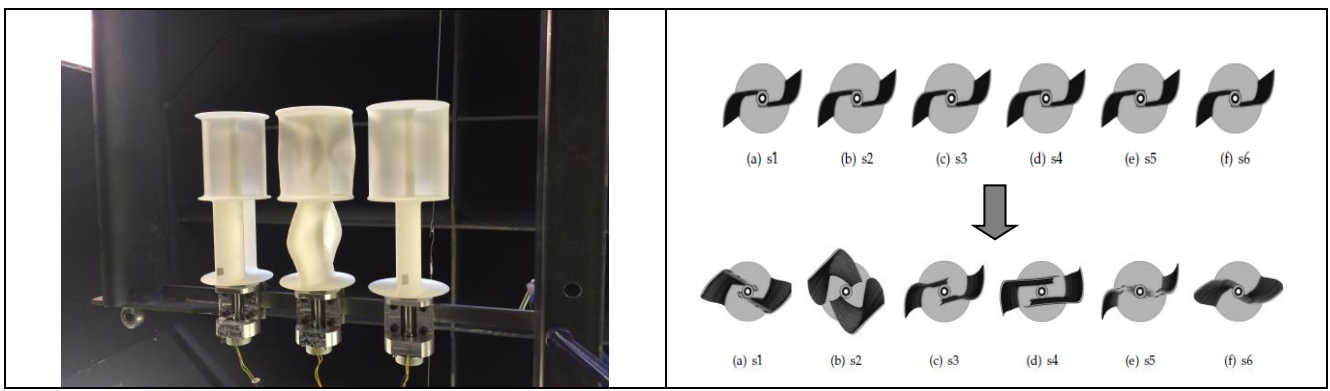

Figure 1: Interacting vertical axis wind turbine design mining.

The approach has so far been developed through its application to the design of arrays of small interacting vertical axis wind turbines [Preen & Bull, 2016; 2017] and cascades of microbial fuel cells [Preen et al., in press]. Evolutionary computing techniques have been used for the optimisation and neural networks for the surrogate modelling. Whilst computational fluid dynamics (CFD) simulations have been successfully applied to the design of wind turbines, they become very computationally expensive as the number of turbines increases. Moreover, various assumptions must be made and accurately modelling the complex wake interactions is an extremely challenging task where different turbulence models can have a dramatic effect on turbine performance; CFD studies have presented significant differences between results even with identical geometric and flow conditions (eg, [Islam et al., 2008]). The same applies to the design of arrays of vertical axis turbines where the aim is to exploit the interactions between the turbines (eg, [Craig et al., 2017]). All such previous work has assumed an array of homogeneous interacting vertical axis turbines; arrays consisted of turbines designed to work as a standalone device. Our initial work used the approach to discover arrays of 2-6 novel heterogeneous vertical axis wind turbines, ie, for the first time turbines were designed to work as part of an array to exploit the interactions (Fig 1, left). Each turbine was designed in parallel: a coevolutionary computation approach was used (eg, [Bull & Fogarty, 1993]), with one design optimising population using its own surrogate model/data per turbine in the array. It was shown possible to exploit the complex inter-turbine turbulent flow conditions to design an array of position specific turbines (Fig 1, right). The starting design for the turbines was based upon a standard commercial design, albeit from a physically larger turbine, and performance in the arrays was typically seen to at least double. Most recently, we have considered the design of cascades of microbial fuel cells (MFC) arranged in series, with the output feedstock fluid of one being used by the next in line. Such cascades have been shown to be more efficient than simply increasing the size of a single MFC. Here the environment of each MFC at any given position in the cascade is potentially very different from that of the others primarily due to the depletion of nutrients in the feed: cascade position specific MFC designs are required. The task therefore contains a living substrate and complex fluid dynamics both within and between the MFC. An approach was adopted where the anode of each MFC was augmented with a 3D printed conductive polymer insert which affected both biofilm shape and nutrient flow. Using cascades of four MFC, each being designed in parallel, it was shown that performance could be increased: ~20% more power and ~100% more power density.

We have recently created examples of the basic aerostat configuration to explore their behaviour and performance. In the smaller, high-speed section of our wind tunnel a fixed position - not buoyant - approximation of two interacting turbines ~0.5m wide has been built (Fig 2, left). Similarly, in a larger section of the wind tunnel, two ~3m wide helium balloons augmented with turbine blades have been assembled (Fig 2, right). Initial testing has confirmed that blades can be attached securely in both cases - despite some collisions and significant deflections in the blimps' surfaces at high wind speeds - by the use of reinforced glass filament strapping tape which also allows quick interchanging of the blades. The fixed turbine position configuration is intended to enable the systematic exploration of the turbine interactions as balloon and blade designs/proportions are varied. That is, this set-up and variants thereof, eg, using rope support, will enable experimentation under relatively controlled conditions along the lines we have previously used to explore interacting vertical axis wind turbines of a similar scale. The larger balloon turbines will enable the subsequent testing of designs and principles identified on the smaller set-up under more realistic (variable/noisy, etc.) conditions and scale (Reynolds number, etc.).

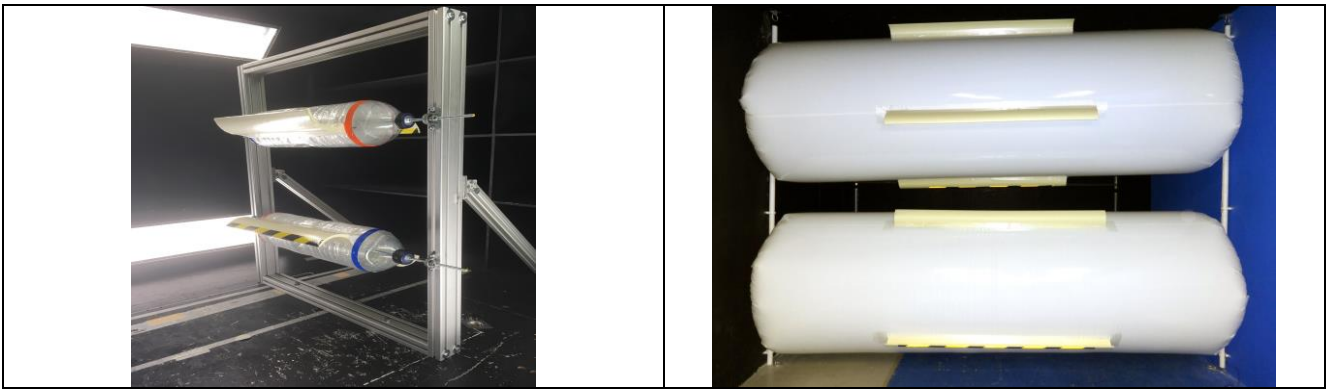

Figure 2: Experimental setups for novel aerostat wind turbine array testing.

For initial results to be present here, using the fixed position aerostats, we have begun by exploring energy generation under different wind conditions, as well as the lift and wake interactions between the turbines through the use of load cells on the end supports of the turbines and the surrounding frame. The basic experimental scenario is very similar to that used in our previous vertical axis wind turbine research: dynamo readings from each turbine (ends) are taken over fixed time periods, for a set of different wind speeds. Experimentation suggests that counter-rotating balloons exhibit more separating lift than co-rotating balloons. The effects of different numbers of blades and blades of different shapes/sizes at different inter-turbine spacings have also been investigated, with initial designs taken/inspired from the vertical axis turbine literature, including our previous work in the area, as well as designs proposed for single turbine aerostats.